\documentclass[letterpaper, 10 pt, conference]{ieeeconf}  % Comment this line out if you need a4paper

\IEEEoverridecommandlockouts                              % This command is only needed if 
\usepackage{graphics} % for pdf, bitmapped graphics files
\usepackage{epsfig} % for postscript graphics files
\usepackage{mathptmx} % assumes new font selection scheme installed
\usepackage{times} % assumes new font selection scheme installed
\usepackage{amsmath} % assumes amsmath package installed
\usepackage{amssymb}  % assumes amsmath package installed
\usepackage{float}
\usepackage{orcidlink}
\usepackage{booktabs}
\usepackage{multirow}

\usepackage{hyperref}
\usepackage{caption}
\usepackage{color}
\usepackage{colortbl}
\usepackage{booktabs}
\usepackage{multirow}
\usepackage{xcolor}
\usepackage{caption}
\usepackage{gensymb}
\usepackage{url}
\usepackage{amsmath} % assumes amsmath package installed
\usepackage{amssymb}  % assumes amsmath package installed
\usepackage{subfigure}
\usepackage{array}
\usepackage{soul}
\usepackage{booktabs}
\usepackage{tcolorbox}
\usepackage{color,xcolor}
\usepackage{multirow}

\usepackage{float}
\usepackage{cite}
\usepackage{algpseudocode}
\usepackage[export]{adjustbox}
\usepackage{balance}
\usepackage{rotating}
\usepackage{graphicx}
\usepackage{hyperref}
\usepackage{wrapfig}
\definecolor{change-red}{RGB}{150,25,25}

\definecolor{mygreen}{RGB}{41,200,51}
\definecolor{myred}{RGB}{160,30,30}
\definecolor{myblue}{RGB}{255,0,0}

\title{\LARGE \bf
TFTrack: A \underline{T}emplate-\underline{F}ree Framework for Efficient 3D Point Cloud \underline{Tracking}
}

\author{Zhaofeng Hu$^{1}$, Sifan Zhou$^{2}$, Jiahao Nie$^{3}$, Ziyu Zhao$^{4}$, Weizi Li$^{5}$, Ci-jyun Liang$^{1}$
\thanks{$^{1}$ Stony Brook University. $^{2}$ Carnegie Mellon University. $^{3}$ Hangzhou Dianzi University $^{4}$ 
Southeast University.
 $^{5}$ University of California, Riverside. }
}
\usepackage{xcolor}

\begin{document}

\maketitle
\thispagestyle{empty}
\pagestyle{empty}

%%%%%%%%%%%%%%%%%%%%%%%%%%%%%%%%%%%%%%%%%%%%%%%%%%%%%%%%%%%%%%%%%%%%%%%%%%%%%%%%
\begin{abstract}

LiDAR-based 3D Single Object Tracking (3D SOT) is critical for robotic perception and navigation and aims to localize dynamic objects across frames in sparse point clouds. Existing methods, rooted in the Siamese tracking paradigm from 2D vision, rely on costly dual-input designs and excessive motion modeling guided by template priors, hindering their efficiency. Our in-depth analysis reveals: (i) the template paradigm is redundant, as the previous bounding box center encodes sufficient historical context; (ii) complex motion modeling is unnecessary, as geometric alignment provides adequate motion priors. Based on the above findings, we propose the first Template-Free Tracking framework (TFTrack). The novel framework eliminates the need for template-search pairings and operates directly on the current frame guided solely by the prior bounding box center and size. We instantiate this paradigm into three variants: TFTrack-Voxel, TFTrack-Pillar, and TFTrack-Point, to explore different 3D representations under a unified framework, ensuring flexibility across sparse and dense scenes. Extensive experiments on KITTI and nuScenes benchmarks show that TFTrack is competitive with leading template-based trackers, while reducing FLOPs by $\sim$50\% and running at $\sim$120 FPS. By simplifying overcomplicated motion-centric designs, TFTrack establishes a new minimalist paradigm for efficient 3D point cloud tracking, paving the way for real-time and resource-efficient deployment in embedded robotic systems, such as autonomous vehicles. The \href{https://github.com/tftrack-anonymous/TFTrack/tree/main}{\textcolor{red}{code}} is available.

% \href{https://github.com/StiphyJay/MVCTrack}{\raisebox{-0.05\height} \ \textbf{\textcolor{red}{{Code}}}} and \href{https://youtu.be/c-OPJ0PvvbA}{\raisebox{-0.05\height} \ \textbf{\textcolor{red}{{video}}}} are available.

\end{abstract}
%
% \begin{keywords}
% Model compression, referring image segmentation, post-training quantization
% \end{keywords}

\section{INTRODUCTION}
LiDAR-based 3D Single Object Tracking (3D SOT) is pivotal for perception and navigation in mobile robots and autonomous vehicles \cite{hu2023aerial,li2023motion,cui2019point,hu2025mvctrack}. It enables real-time localization of dynamic objects across frames using sparse point clouds. This process estimates the target's four degree-of-freedom~(DoF) pose (position $(x,y,z)$ and yaw $\theta$) under the common assumption that objects are upright, ground-supported, and primarily rotate around the vertical axis. For instance, a person-following robot relies on 3D SOT to reliably track and estimate the person’s 3D position relative to the robot in crowded environments, which is essential for effective following control. Similarly, during UAV landing, the UAV determines its distance to the target and poses for a secure landing. These capabilities support critical tasks like path planning \cite{han2021planning}, obstacle avoidance \cite{snyder2023online}, and interactive decision-making \cite{unhelkar2020semi}.

\begin{figure}[t]
    \centering
    \includegraphics[width=\linewidth]{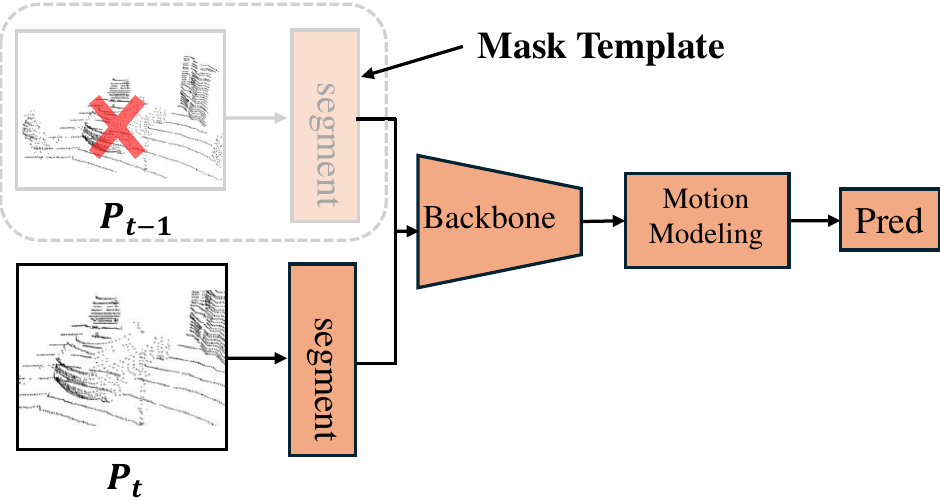}
    \caption{Illustration of the template-based tracking architecture. The network extracts features from both the template $\mathbf{P}_{t-1}$ and search $\mathbf{P}_t$ frames through a shared backbone, and performs motion modeling for prediction. Masking the template simulates a degraded reference and reveals redundancy.} 
    \label{fig:template_architecture}
    \vspace{-7mm}
\end{figure}

Mainstream 3D SOT methods \cite{p2b, 3dsiamrpn, bat, ptt, cxtrack, mbptrack,pillartrack} are derived from 2D visual tracking paradigms\cite{kcf,eco,siamrpn,siamfc} and predominantly rely on siamese-based, appearance-matching architectures. These methods employ dual-input backbones to extract features from a template and search region, followed by a correlation module to regress target poses. Despite their accuracy, these dual-branch siamese trackers incur substantial computational overhead through costly template-search pairings, neglecting motion cues critical for sparse or occluded point clouds. Moreover, the inherently sparse and textureless nature of LiDAR point clouds significantly weakens their representation capability under challenging conditions such as occlusions, appearance variations, and viewpoint changes, consequently limiting their robustness and generalization in dynamic environments. Recent motion-centric paradigms have emerged as robust alternatives \cite{M2Track,m2track++,li2024flowtrack}. By reformulating tracking as relative motion estimation across consecutive frames, these methods reduce reliance on appearance, improving performance in sparse environments. However, they often overcomplicate motion modeling with multi-frame inputs and uniform motion assumptions, failing to distinguish foreground dynamics from static backgrounds. Background noise complicates motion estimation, necessitating refinement modules, as in M$^2$Track \cite{M2Track}. Such multi-stage pipelines and complex motion modeling hinder efficiency, especially for resource-constrained robotic platforms.

\vspace{-1mm}
\begin{figure}[t]
\begin{minipage}{0.9\linewidth}
    \centering
    \includegraphics[width=\linewidth, trim=0 7 0 15, clip]{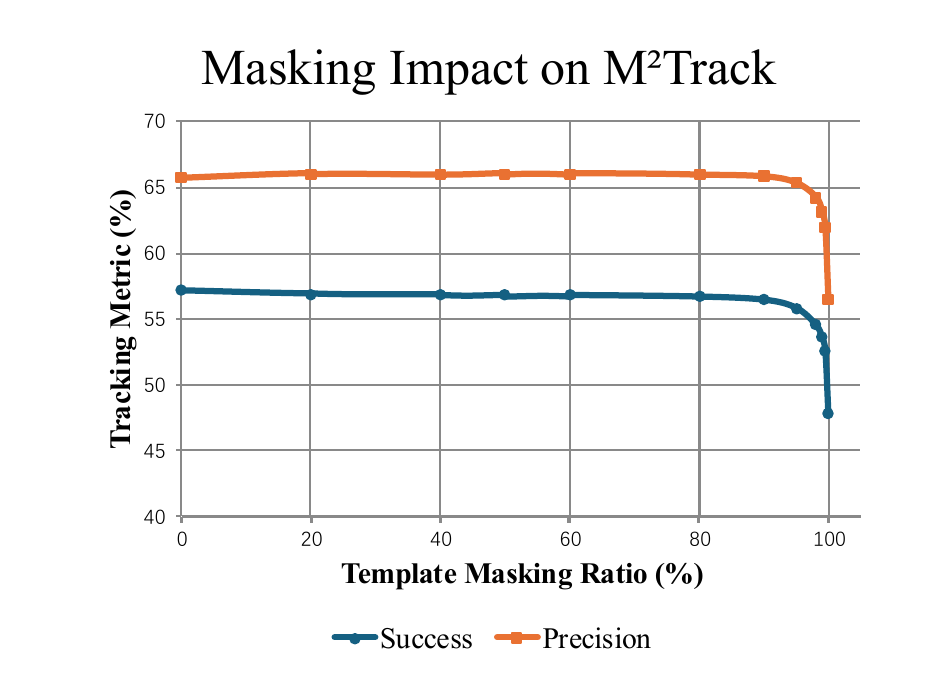}
    \vspace{-8mm}
    \caption{Performance drop of M$^2$Track under increasing template masking ratios on nuScenes.}
    \label{fig:template_structure}
\end{minipage}

\vspace{1mm}

\begin{minipage}{\linewidth}
    \centering
    \captionof{table}{Numerical results of M$^2$Track under different template masking ratios on nuScenes. 
See Fig.~\ref{fig:template_structure} for the performance curve.}
    \vspace{1mm}
    \begin{adjustbox}{max width=0.6\linewidth}
        \begin{tabular}{c|cc}
            \toprule
            Mask Ratio & Success (\%) & Precision (\%) \\
            \midrule
            0\% (baseline)  & 57.21 & 65.73 \\
            20\%            & 56.89 & 66.03 \\
            40\%            & 56.86 & 66.01 \\
            60\%            & 56.86 & 66.04 \\
            80\%            & 56.74 & 66.00 \\
            100\%           & 47.86 & 56.58 \\
            \bottomrule
        \end{tabular}
    \end{adjustbox}
    \vspace{-3mm}
    \label{tab:masking_ablation}
\end{minipage}
\vspace{-3mm}
\end{figure}

Overall, the current 3D SOT trackers, whether appearance matching or motion-centric models, face two fundamental challenges: \textbf{(1) Computational Inefficiency:} These methods process two point cloud sequences in parallel, leading to duplicated feature extraction. This doubles both FLOPs and memory footprint, making them ill-suited for the strict power and latency constraints of embedded robotic hardware. \textbf{(2) Questionable Necessity of Full Template Information:} The reliance on the template frame (previous) raises doubts about its true contribution. Specifically, it is unclear how much additional information the full template provides beyond the compact box prior, especially given the associated computational cost. To quantify how much information the template frame actually contributes, we perform a \emph{mask-out ablation experiment}, as shown in Fig.~\ref{fig:template_architecture} and Fig.~\ref{fig:template_structure}. We fed a state-of-the-art motion-centric tracker \cite{M2Track} with the current frame while zeroing out the template input with different mask ratios. The results indicate that the average success drops by less than 1\% until the mask ratio reaches 80\%, as shown in Table~\ref{tab:masking_ablation}. This counterintuitive observation reveals: \textbf{(1) in the evaluated motion-centric setting, the prevailing Siamese dual-input paradigm introduces substantial redundancy in 3D SOT}, as the performance remains largely intact without the template input; \textbf{(2) the template point cloud is rarely indispensable for object tracking}, given the minimal performance drop when it is removed. These findings show that the template frame may not always be necessary in representative 3D SOT methods.

To this end, we present the \textbf{first \underline{t}emplate-\underline{f}ree} framework for 3D single object \underline{track}ing (TFTrack). By processing only the current point cloud with a siamese-free backbone and initializing with the prior bounding box center, TFTrack treats motion estimation as a lightweight \emph{geometric transformation}. This design eliminates the template branch and the cross-frame matching head trims inference FLOPs by roughly 50\% while maintaining competitive accuracy We further instantiate TFTrack in three variant trackers: \textbf{TFTrack-point}, \textbf{TFTrack-voxel}, and \textbf{TFTrack-pillar}, balancing speed and performance across different 3D representations. All variants forgo the overhead of Siamese matching and achieve highly real-time speed ($\sim$120~FPS on a single NVIDIA 3090 GPU), making TFTrack suitable for resource-constrained robotic systems. Our main contributions are summarized as:
% \vspace{-2mm}

\begin{itemize}
\item \textbf{Revelation of Siamese Redundancy in 3D SOT}: Through extensive ablations on representative motion-centric Siamese 3D SOT, we reveal substantial redundancy in dense template point clouds, exposing inherent redundancy in vision-inherited Siamese networks for point cloud. 
  % \vspace{-1mm}
  \item \textbf{TFTrack: Template-Free 3D SOT Paradigm}: We propose TFTrack, the first template-free 3D tracker that reformulates 3D SOT as a lightweight geometric transformation around the prior bounding box center.
  % \vspace{-1mm}
  \item \textbf{Unified and Efficient 3D SOT Framework}: We demonstrate the effectiveness of the unified TFTrack framework with three variants (point, voxel, and pillar) on KITTI and nuScenes datasets, achieving competitive accuracy at up to $\sim$120 FPS, especially suitable for resource-constrained robotic platforms.
\end{itemize}

\begin{figure*}[ht]
    \centering
    % \hspace*{-0.12\linewidth}
    \includegraphics[width=0.85\linewidth, trim=0 10 0 0, clip]{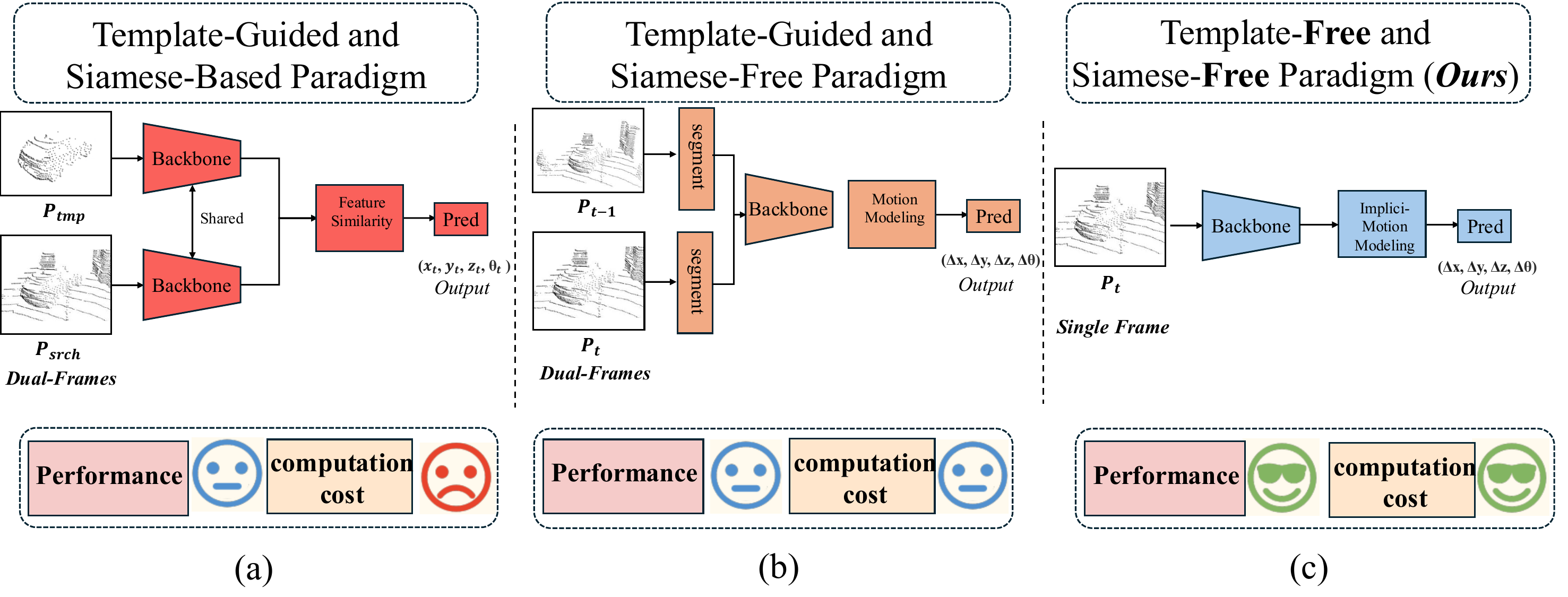}
    % \vspace{-2mm}
    \caption{Comparison of three different 3D SOT architectures. \textbf{(a)} \textit{Template-Guided and Siamese-Based Paradigms} \cite{p2b,osp2b,ptt,bat,v2b} leverage a siamese architecture for template points and search points, followed by similarity matching, yielding low efficiency. \textbf{(b)} \textit{Template-Guided and Siamese-Free Paradigms} \cite{M2Track,m2track++} use previous and current points to model the relative motion, with sub-optimal speed and accuracy. \textbf{(c)} \textit{Template-Free and Siamese-Free Paradigm (Ours)}: Our design removes reliance on template point clouds and Siamese backbones. It directly takes the current frame as input, implicitly encoding prior motion context in an effective one-shot pipeline, achieving high efficiency and accuracy.}
    \label{fig:compare}
    \vspace{-4mm}
\end{figure*}

\vspace{-2mm}
\section{RELATED WORK}
\label{sec:related}
\noindent\textbf{Appearance Matching for 3D SOT.} Early 3D SOT methods formulate tracking as cross-frame appearance matching through Siamese architectures. They use Siamese networks for feature extraction and reveal local tracking cues based on similarity matching. While SC3D \cite{sc3d} pioneered template-based similarity learning, its non-end-to-end design limits candidate generation. Subsequent methods such as P2B \cite{p2b} and 3D-SiamRPN \cite{3dsiamrpn} employ target-specific feature matching and region proposals, but template dependence introduces computational overhead. PTT \cite{ptt} introduced a Point-Track-Transformer to weight critical point cloud features for enhanced appearance-based matching. In addition, PTTR \cite{pttr} and STNet \cite{stnet} explored different attention mechanisms to improve feature correlation. Recently, MoCUT \cite{cutrack} and TrackAny3D \cite{wang2025trackany3d} introduced unified appearance-based tracking across object categories. Despite their advances, template-based methods incur high computational costs and limit real-time deployment because of their dual-input designs. Moreover, their reliance on appearance breaks down in sparse and occluded point clouds, overlooking motion cues sufficient for 3D tracking.

\noindent\textbf{Motion Modeling for 3D SOT.} The limitations of Siamese-based methods in handling sparse, texture-less point clouds have prompted efforts \cite{M2Track,m2track++,dmt,li2024flowtrack,bevtrack} to model inter-frame motion. M$^2$Track \cite{M2Track} introduced a motion-centric framework that leverages geometric consistency across frames, reducing reliance on appearance cues. M$^2$Track++ \cite{m2track++} extended this framework to a semi-supervised setting to further enhance performance. Despite improved robustness, existing methods still explicitly rely on complex multi-frame motion modeling pipelines that introduce computational redundancy. By neglecting simple geometric alignments, such as those derived from prior box centers, they limit efficiency and robustness, motivating our template-free approach. 

\section{METHOD}
\subsection{Overview}

% \label{defi}
% \noindent\textbf{Task Definition.} 
We define the task of LiDAR-based 3D Single Object Tracking (3D SOT) as follows: given a point‐cloud sequence $\{{P}_t\}_{t=1}^{T}$ with ${P}_t\in\mathbb{R}^{N\times3}$ and an initial box 
${B}_1 = (x_1,y_1,z_1,w,h,l,\theta_1)\in\mathbb{R}^7$ in the first frame, where $(x,y,z)$ and $(w,h,l)$ denote the center and size, $\theta$ is the rotation angle around vertical $z$-axis. The goal of 3D SOT is to predict \({B}_t=(x_t,y_t,z_t,w,h,l,\theta_t)\) for each subsequent frame \(t=2,\dots,T\). Following \cite{mbptrack,ptt,bat}, for both rigid and non-rigid objects, since the size of the target \((w,h,l)\) remains approximately unchanged in 3D SOT across all frames, tracking reduces to regressing the 4-DoF parameters \((x_t,y_t,z_t,\theta_t)\in\mathbb{R}^4\) to represent ${B}_{t}$ at every frame.

\subsection{Revisit the Template-based Tracking Paradigm}
\label{siamese_revist}

Mainstream 3D trackers \cite{ptt,cxtrack,mbptrack,M2Track,osp2b,m2track++,wang2023correlation} adopt the \emph{Siamese} paradigm with dual-input (i.e., template-search pairs), processing two point-cloud inputs via weight-shared backbones and a cross-frame prediction head. As shown in Fig~\ref{fig:compare}(a)-(b), there are two dominant variants: 
\begin{itemize}
    % \vspace{-1mm}
    \item \textbf{Appearance-based (template–search) tracking} \cite{p2b,ptt,osp2b} selects the point cloud from the first frame based on the ground truth box as template ${P}^{\text{t}}$\footnote{Some methods \cite{cxtrack,mbptrack} also adopted the previous frame as template.} and crop points around the previous prediction ${B}_{t-1}$ as search region ${P}^{\text{s}}$. After the siamese-based backbone, an appearance matching module ${F}_{\text{app}}$ measures feature similarity to regress the 4-DoF parameters $(x_t, y_t, z_t, \theta_t)$.
    % \vspace{-1mm}
    \item \textbf{Motion-centric (inter-frame) tracking} \cite{M2Track, m2track++} segments foreground points from dual frames $({P}^{\text{crop}}_{t-1}, {P}^{\text{crop}}_{t})$ and models motion to regress the relative offset $(\Delta x, \Delta y, \Delta z, \Delta \theta)$, then applies rigid-body transformation to the previous prediction \({B}_{t-1}\) to obtain $(x_t, y_t, z_t, \theta_t)$. %The final target bounding box \((x_t, y_t, z_t, \theta_t)\) is then obtained by applying a rigid-body transformation to the previous prediction \({B}_{t-1}\). %${F}_{\text{mot}}$ to get the motion-cues, after the motion head ${H}_{\text{mot}}$ regresses the relative offset $(\Delta x, \Delta y, \Delta z, \Delta \theta)$. The final target bounding box \((x_t, y_t, z_t, \theta_t)\) is then obtained by applying a rigid-body transformation to the previous prediction \({B}_{t-1}\).
\end{itemize}
These process can be formulated as follows,

\noindent\textbf{Template-based Appearance Matching: }
\begin{align}
\label{eq:siamese_app}
(x_t, y_t, z_t, \theta_t) = 
{F}_{\text{app}}\left({P}^{\text{tmp}}, {P}^{\text{srch}}_t\right)
\end{align}

% \vspace{-2mm}

\noindent\textbf{Template-based Motion-Centric: } 
\begin{align}
\label{eq:siamese_mot}
(x_t, y_t, z_t, \theta_t) =
{F}_r\Big( {F}_m\big( {B}_{t-1},\,
{F}_s( {P}^{\text{crop}}_{t-1},\, {P}^{\text{crop}}_t ) \big) \Big)
\end{align}

% \begin{multline}
% \label{eq:siamese_app}
% \textbf{Template-based Appearance Matching: } \\
% (x_t, y_t, z_t, \theta_t) = {F}_{\text{app}}\left({P}^{\text{tmp}}, {P}^{\text{srch}}_t\right)
% \end{multline}
% \begin{multline}
% \label{eq:siamese_mot}
% \textbf{Template-based Motion-Centric: } \\
% (x_t, y_t, z_t, \theta_t) =
% {F}_r\Big( {F}_m\big( {B}_{t-1},\,
% {F}_s( {P}^{\text{crop}}_{t-1},\, {P}^{\text{crop}}_t ) \big) \Big)
% \end{multline}

\noindent where \({F}_{\text{app}}\) denotes the appearance matching module, and ${F}_s,{F}_m,{F}_r$ denote the segmentation module, motion inference module, and motion refine module \cite{M2Track,m2track++}. Despite their effectiveness, these methods double the input size, introduce computational redundancy in cross-frame matching, and complicate motion modeling, neglecting the sufficient geometric alignments derived from prior boxes. These limitations necessitate our \emph{template-free}, single-frame approach.

% While effective, both formulations necessarily double the input size, require cross-frame feature matching, and heavily tie performance to the quality of intermediate modules.  
% These limitations motivate our forthcoming \emph{Siamese-free}, single-frame formulation.

% Both branches fit different Siamese formulations:
% \begin{equation}
% \label{eq:siamese_app_new}
% (x_t, y_t, z_t, \theta_t) = 
% {H}_{\text{app}}\!\left({F}_{\text{app}}\left({F}_{\text{seg}}({P}^{\text{tmp}}), {P}^{\text{srch}}_t\right)\right).
% \end{equation}
% \begin{equation}
% \begin{aligned}
% (\Delta x, \Delta y, \Delta z, \Delta \theta) &= 
% {H}_{\text{mot}}\!\left({F}_{\text{mot}}\left({F}_{\text{seg}}\left({P}^{\text{crop}}_{t-1}, {P}^{\text{crop}}_t\right)\right),\, {B}_{t-1}\right), \\
% (x_t, y_t, z_t, \theta_t) &= 
% \text{Transform}\left({B}_{t-1}, (\Delta x, \Delta y, \Delta z, \Delta \theta)\right).
% \end{aligned}
% \end{equation}

% where ${F}_{\text{app}}$ and ${F}_{\text{mot}}$ denote the appearance matching modules and motion modeling modules, ${F}_{\text{seg}}$ denotes the foreground segmentation module, and ${H}_{\text{app}}$, ${H}_{\text{mot}}$ are the final prediction heads.  

\begin{figure*}[ht]
\centering
\includegraphics[width=0.99\textwidth]{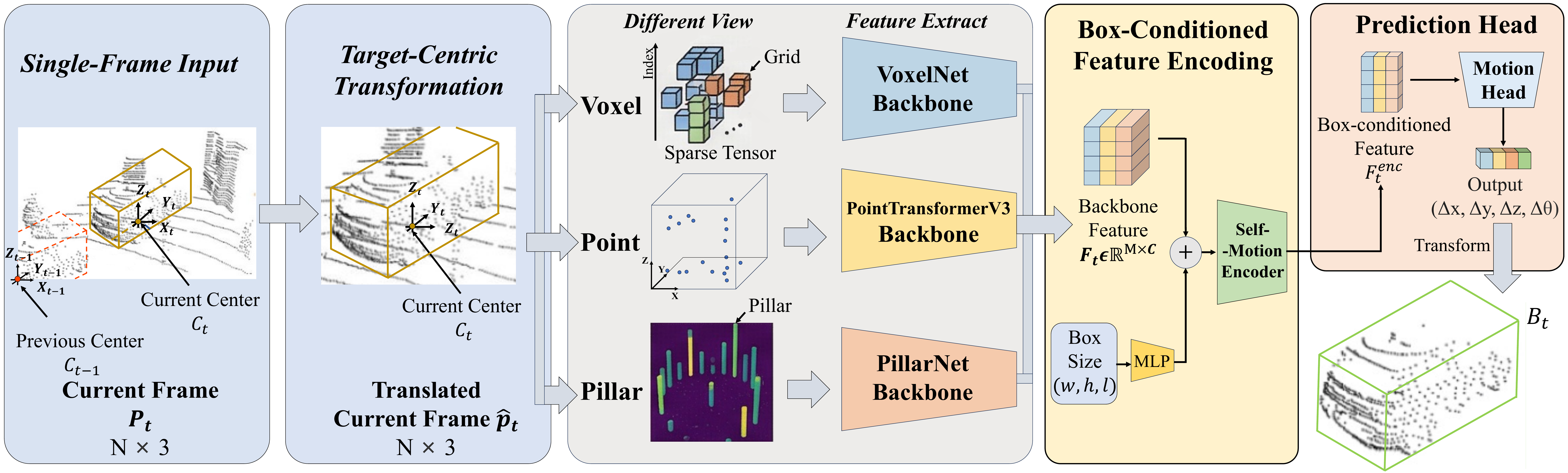}
\vspace{-1mm}
\caption{Our proposed template-free tracking framework TFTrack. Given a single-frame point-cloud input, TFTrack supports point-, voxel-, and pillar-based backbones, to extract geometric features. These features are processed by Box-Conditioned Feature Encoding (BCFE) and fed into a motion head to predict relative offset, enabling efficient 3D SOT without a Siamese architecture and template input.}
\label{fig:framework}
\vspace{-6mm}
\end{figure*}

\subsection{Unifying Template-Free Framework for 3D SOT}
\label{siamfree}

As shown in Fig~\ref{fig:compare} (c), unlike existing methods, our TFTrack pioneers a template-free framework for 3D SOT, addressing the computational redundancy of template-based trackers \cite{p2b,M2Track}. Motivated by the ablation results showing minimal performance loss without templates, we leverage the prior bounding box center as a lightweight motion prior, eliminating dual-frame inputs. We reformulate the paradigm of 3D SOT as a template-free prediction:

\noindent\textbf{Template-free Motion Modeling:}
% \vspace{-2mm}
\begin{align}
\label{eq:single_frame_motion}
\begin{aligned}
% \hat{{P}}_{t}
% &= {P}_{t} - \mathbf{c}\bigl({B}_{t-1}\bigr), \\[-3pt]
(\Delta x, \Delta y, \Delta z, \Delta \theta) = {F}_{\mathit{sf}}\bigl({P}_{t}^{\mathrm{crop}}\bigr),
\end{aligned}
\end{align}
where ${P}_{t}^{\mathrm{crop}}$ is the point cloud from current frame, and ${F}_{\mathit{sf}}$ denotes our template-free and siamese-free paradigm. This design eliminates the need for template points, cross-frame feature matching and multi-frame motion modeling, reducing both computation and memory. As shown in Fig.~\ref{fig:framework}, our \textbf{TFTrack} framework unifies three variants: \textbf{TFTrack-point}, \textbf{TFTrack-voxel}, and \textbf{TFTrack-pillar}, which use point, voxel, and pillar representations for different application scenes, respectively. Each variant employs a lightweight backbone to extract geometric features, followed by a Box-Conditioned Feature Encoding module and a prediction head to regress the final position.

\subsubsection{\textbf{Target-Centric Transformation}}
Given the current frame point cloud ${P}_t = \{\mathbf{p}_i\}_{i=1}^{N}$ and the previous bounding box center $\mathbf{c}_{t-1}$, we leverage \(\mathbf{c}_{t-1}\) as a location prior, defining \(\hat{{P}}_t = \{\mathbf{p}_i - \mathbf{c}_{t-1}\}_{i=1}^N\). This reformulates tracking as predicting a 4-DoF motion rigid transformation that aligns \(\hat{{P}}_t\) to the target’s true location in frame $t$. The centering operation embeds historical motion context into spatial coordinates, eliminating the need for explicit template matching while preserving geometric continuity across frames.

\subsubsection{\textbf{Different Point Cloud Views}}
To validate the effectiveness, geometric fidelity, and compatibility of our proposed TFTrack framework, we implement three point cloud representations:  \emph{voxel}, \emph{point} and \emph{pillar}. Given the input point cloud $\hat{{P}}_{t}\in \mathbb{R}^{N \times C}$, we define three representations:
\begin{itemize}
    \item \textbf{Voxel:} The scene is partitioned into a 3D grid along X, Y, Z axes, where z-axis partitioning complicates fitting a tensor $[B, X, Y, Z, C]$ into memory. Dense grids employ a structure of size $(s_x, s_y, s_z)$, forming \(H \times W \times D\) cells. Sparse encoding processes non-empty voxels with a separate $[M, 4]$ matrix, which stores their indices. We therefore adopt a sparse format for voxel.

    \item \textbf{Point:} The point cloud $\hat{P}_t$ is represented as a feature matrix of dimensions $N\times3$, corresponding to the Cartesian coordinates $(x, y, z)$ for each of the $N$ points. This representation treats the input as an unordered set, retaining raw geometric information.

    \item \textbf{Pillar:} From a top-down view~\cite{zhou2025fastpillars,zhou2025pillarhist}, \(\hat{{P}}_{t}\) is projected to an \(xy\)-grid, forming a 2D pseudo-image of vertical columns. Dense representations use a feature tensor $[B, X, Y, C]$, where $B$ is the batch size, $(X,Y)$ denote pillar counts. Sparse representations encode non-empty pillars into a matrix $[M, C]$, with an index matrix $[M, 3]$ for batch and spatial indices of non-empty pillars. We adopt a sparse format for non-empty pillars into a tensor, enabling high-speed processing.

\end{itemize}
% \vspace{-1mm}
By processing only the current frame, all variants halve FLOPs compared to Siamese or motion-centric backbone designs, while leveraging \(\mathbf{c}_{t-1}\) for lightweight geometric alignment, thus redefining efficient and robust 3D tracking.

\subsubsection{\textbf{Siamese-free Backbone}} We adopt a template-free architecture that encodes only the current point cloud $\hat{{P}}_t = \{\hat{\mathbf{p}}_i\}_{i=1}^N$. This simple shift enables the model to implicitly capture motion context without relying on dual-frame inputs or cross-frame feature matching. A siamese-free backbone $\Phi: \mathbb{R}^{N \times 3} \rightarrow \mathbb{R}^{M \times C}$, where $M$ denotes the number of feature tokens (e.g., voxels, points, or pillars), $C$ denotes the backbone-dependent feature dimension, is applied to extract features in a forward pass: $\mathbf{F}_t = \Phi(\hat{{P}}_t) \in \mathbb{R}^{M \times C}$. Based on the input representation, the backbone can be instantiated as PointTransformerV3 \cite{ptv3} for point-based, VoxelNet \cite{voxelnet} for voxel-based, and PillarNet \cite{shi2022pillarnet} for pillar-based. We compare other backbones in Table~\ref{table_backbone}. Despite architectural differences, all variants adopt the same siamese-free paradigm.

\begin{table*}[t]
\vspace{-1mm}
\centering
\caption{Comparisons with state-of-the-art methods on KITTI dataset. The upper and lower parts include dual-frame and single-frame trackers, respectively. \textit{Success} / \textit{Precision} are used for evaluation. $\dagger$ denotes reimplementation based on official code. Mean is frame-weighted.}

    \resizebox{0.9\textwidth}{!}{
    \normalsize
    \begin{tabular}{c|cc|c|cccc|cc}
          \toprule[0.4mm]
            && & Mean& Car & Pedestrian &  Van & Cyclist &  \\
           \multirow{-2}{*}{Paradigm}&\multirow{-2}{*}{Tracker} & \multirow{-2}{*}{Source} & (14,068) & (6,424)&(6,088) & (1,248) & (308) & \multirow{-2}{*}{FPS} & \multirow{-2}{*}{Device} \\
          \midrule
          \multirow{15}{*}{Template-based}&SC3D~\cite{sc3d}& CVPR'19 & 31.2 / 48.5 &41.3 / 57.9   & 18.2 / 37.8 & 40.4 / 47.0 & 41.5 / 70.4   &2&GTX 1080Ti \\
          &P2B~\cite{p2b}& CVPR'20& 42.4 / 60.0 & 56.2 / 72.8 & 28.7 / 49.6 & 40.8 / 48.4 & 32.1 / 44.7   &40&GTX 1080Ti\\ %& 4.30 G\\ %\multirow{18}{*}{Matching}
        &PTT~\cite{ptt-journal}&TMM'23 & 55.1 / 74.2 &67.8 / 81.8& 44.9 / 72.0 &43.6 / 52.5& 37.2 / 47.3 & 40&GTX 1080Ti \\ %& - \\
        &LTTR~\cite{lttr}&BMVC'21& 48.7 / 65.8 & 65.0 / 77.1 &33.2 / 56.8 & 35.8 / 45.6 &66.2 / 89.9 & 23 &GTX 1080Ti \\ %& - \\
        % &MLVSNet~\cite{mlvsnet}&ICCV'21 & 45.7 / 66.6 &56.0 / 74.0   & 34.1 / 61.1 & 52.0 / 61.4 & 34.4 / 44.5  & 70 &GTX 1080Ti \\% &  -\\
         &BAT~\cite{bat}&ICCV'21& 51.2 / 72.8 & 60.5 / 77.7 &42.1 / 70.1 & 52.4 / 67.0 &33.7 / 45.4 & 57 &RTX 2080  \\%& 2.77 G\\
        % &V2B~\cite{v2b}&NeurIPS'21 & 58.4 / 75.2 &70.5 / 81.3   & 48.3 / 73.5 & 50.1 / 58.0 & 40.8 / 49.7   &37&TITAN RTX \\ %5.57 G\\
        &PTTR~\cite{pttr}& CVPR'22& 57.9 / 78.2 &65.2 / 77.4   & 50.9 / 81.6 & 52.5 / 61.8 & 65.1 / 90.5  &50 &Tesla V100 \\ %& 2.61 G\\
        &STNet~\cite{stnet} & ECCV'22 & 61.3 / 80.1 &72.1 / 84.0 &49.9 / 77.2& 58.0 / 70.6& 73.5 / 93.7& 35 &TITAN RTX \\ %& 3.14 G\\
        % &CMT~\cite{cmt} & ECCV'22 &59.4 / 77.6 &70.5 / 81.9 &49.1 / 75.5& 54.1 / 64.1& 55.1 / 82.4   & 32&GTX 1080Ti \\ %& -\\
        &GLT-T~\cite{glt}&AAAI'23 & 60.1 / 79.3 &68.2 / 82.1   & 52.4 / 78.8 & 52.6 / 62.9 & 68.9 / 92.1  &30 &GTX 1080Ti \\ %& 3.87 G\\
        &OSP2B~\cite{osp2b}& IJCAI'23& 60.5 / 82.3 &67.5 / 82.3   & 53.6 / 85.1 & 56.3 / 66.2 & 65.6 / 90.5   & 34& GTX 1080Ti \\ %& 2.57 G\\
        &CXTrack~\cite{cxtrack} & CVPR'23& 67.5 / 85.3 & 69.1 / 81.6 &67.0 / 91.5& 60.0 / 71.8& 74.2 / 94.3& 34&RTX 3090 \\ %& 4.63 G\\
        &MBPTrack~\cite{mbptrack} & ICCV'23 & 70.3 / 87.9 & 73.4 / 84.8 &68.6 / 93.9 &61.3 / 72.7& 76.7 / 94.3  & 50&RTX 3090  \\ %& 2.88 G\\
         &M$^2$Track~\cite{M2Track} & CVPR'22& 62.9 / 83.4 &65.5 / 80.8   & 61.5 / 88.2 & 53.8 / 70.7 & 73.2 / 93.5  &57 &Tesla V100 \\ %& 2.54 G\\
          &M$^2$Track++~\cite{m2track++} & TPAMI'24& 66.5 / 85.2 &71.1 / 82.7  & 61.8 / 88.7 & 62.8 / 78.5 & 75.9 / 94.0   &57& Tesla V100 \\ %& 2.54 G\\
          &MoCUT~\cite{cutrack}&ICLR'24& 65.8 / 85.0 &67.6 / 80.5 &63.3 / 90.0 &64.5 / 78.8& 76.7 / 94.2 &48& RTX 3070Ti \\ 
    & PillarTrack~\cite{pillartrack} & arXiv'24 & 66.8 / 83.7 
    & 74.2 / 85.1 
    & 59.7 / 84.7
    & 61.0 / 69.2 
    & 78.0 / 95.0 & 27 & RTX 3080 \\
            & BEVTrack$^\dagger$~\cite{bevtrack} & IJCAI'25 &
            71.8 / 89.2 &
            74.9 / 86.5 &
            69.5 / 94.3 &
            66.0 / 77.2 &
            77.0 / 94.7 &
            108 & RTX 3090 \\

          &TrackAny3D~\cite{wang2025trackany3d}&ICCV'25& 67.1 / 85.4 &73.4 / 85.2 &59.6 / 85.6  &70.0 / 82.8 & 74.7 / 94.0 &28& RTX 3090\\
          %& 3.27 G\\
         % &P2P-voxel~\cite{P2P}&Arxiv'24& \textbf{71.7} / \underline{89.4} &73.6 / \underline{85.7} &\textbf{69.6} / 94.0 & \textbf{70.3} / \textbf{83.9} & 75.5 / 94.6 &71& RTX 3090 & -\\
         \midrule
         \rowcolor{gray!18}\multirow{3}{*}{Template-free}&\textbf{TFTrack-Voxel} & \textbf{Ours}& 64.0 / 81.6 & 67.3 / 77.7 & 62.1 / 88.5 & 52.8 / 64.3 &  76.1 / 94.4& \textbf{92} & RTX 3090 \\ %& \textbf{1.44 G}\\
        \rowcolor{gray!18}&\textbf{TFTrack-Point} & \textbf{Ours}& 64.4 / 83.7 & 70.0 / 81.1 & 57.7 / 86.4& 66.8 / 81.9& 69.5 / 92.3 &\textbf{121} & RTX 3090 \\ %& \textbf{1.44 G}\\
        \rowcolor{gray!18}&\textbf{TFTrack-Pillar} & \textbf{Ours}& 60.8 / 79.0 & 66.1 / 77.1 &56.4 / 83.5 & 52.1 / 62.9 & 72.3 / 94.1  &\textbf{95} & RTX 3090 \\ %& \textbf{1.44 G}\\
          \bottomrule[0.4mm]
    \end{tabular}  
}

\label{tab:kitti_results}
\vspace{-4mm}
\end{table*}

\subsubsection{\textbf{Box-Conditioned Feature Encoding}}
\label{sec:bcfe}

After target-centric re-centering with the previous bounding-box center, the current-frame features are represented in an aligned local coordinate system. The previous box size provides an object-aware geometric prior, allowing the network to interpret current-frame features relative to the expected target scale and shape. Given a backbone feature $\mathbf{F}_t \in \mathbb{R}^{M \times C}$, we compute a box-size embedding $\mathbf{s}=\mathrm{MLP}(w,h,l)\in\mathbb{R}^{C}$ and concatenate it to each token as $\tilde{\mathbf{F}}_t=[\mathbf{F}_t \| \mathbf{1}_M\mathbf{s}^{\top}]\in\mathbb{R}^{M\times 2C}$, where $\mathbf{1}_M$ is an all-ones column vector.

The box-conditioned feature is processed by a lightweight encoder $\Psi$ to obtain $\mathbf{F}^{\mathrm{enc}}_t=\Psi(\tilde{\mathbf{F}}_t)$. For point features, $\Psi$ is implemented with point-wise feature-mixing layers; for voxel and pillar features, it uses convolutional blocks over BEV maps to capture local spatial patterns. A domain-specific readout head $R(\cdot)$ then produces the final representation $\mathbf{g}_t=R(\mathbf{F}^{\mathrm{enc}}_t)$ for offset regression. Thus, Box-Conditioned Feature Encoding (BCFE) provides scale- and shape-aware feature conditioning for target localization in the aligned current frame.

\subsubsection{\textbf{Prediction Head and Training Loss}}
A two–layer MLP \({H}:\mathbb{R}^{C}\!\longrightarrow\mathbb{R}^{4}\) predicts the 4–DoF relative offset:
\vspace{-1.8mm}
\begin{equation}
  \begin{aligned}
    \bigl(\Delta x, \Delta y, \Delta z, \Delta\theta\bigr)
    &= H\bigl(\mathbf{g}_t\bigr), \\
    (x_t, y_t, z_t, \theta_t)
    &= \mathrm{Transform}\bigl({B}_{t-1}, (\Delta x, \Delta y, \Delta z, \Delta\theta)\bigr).
  \end{aligned}
  \label{eq:regress_and_update}
\end{equation}
\vspace{-2mm}

At test time, the predicted bounding box is recovered via rigid body transformation, as shown in Eq.~\ref{eq:regress_and_update}. Because the head consumes only a single global descriptor aggregated from the backbone features, its parameter count is identical across the voxel, pillar, and point variants (\(<\!0.1\%\) of the total), keeping the prediction-head capacity consistent across variants. To train the motion head, we adopt residual log-likelihood estimation (RLE) loss~\cite{rle}, which models the motion residual distribution using a flow-based density estimator:
\vspace{-1mm}
\begin{equation}
{L}_{\text{RLE}} = - \log p_\phi(\mathbf{m}_t^{\text{gt}} - \mathbf{m}_t^{\text{pred}}),
\end{equation}
where $p_\phi(\cdot)$ is parameterized by a conditional normalizing flow, and $\mathbf{m}_t^{\text{gt}}, \mathbf{m}_t^{\text{pred}}$ denote the ground truth and predicted motion vectors, respectively.

\section{EXPERIMENTS}
% \paragraph{\textbf{Implementation Details}} 
% The extended range is defined as [(-4.8,4.8),(-4.8,4.8),(-1.5,1.5)] and [(-1.92,1.92),(-1.92,1.92),(-1.5,1.5)] for cars and pedestrian, respectively. After the backbone, the spatial resolution is down-sampled to $16\times 16\times 128$. We follow the common setup~\cite{p2b,M2Track} and conduct extensive experiments on KITTI~\cite{kitti}, nuScenes~\cite{nuScenes}. The evaluation metrics is followed the common setup~\cite{ptt,ptt-journal,M2Track,m2track++,cxtrack} to report Success and Precision based on one pass evaluation (OPE)~\cite{otb2013,kristan2016novel}. Specifically, nuScenes~\cite{nuScenes} dataset present additional challenges over KITTI due to higher scene complexity and larger-scale annotations. See the supplementary materials for more details.
\subsection{Implementation Details}
We follow the LiDAR-based 3D SOT experiments \cite{p2b,M2Track} to evaluate TFTrack. For example, the extended spatial range is defined as $[(-4.8, 4.8), (-4.8, 4.8), (-1.5, 1.5)]$ for cars and $[(-1.92, 1.92), (-1.92, 1.92), (-1.5, 1.5)]$ for pedestrians, reflecting their different physical scales. Following backbone processing, the spatial resolution is downsampled to $16 \times 16 \times 128$. We conduct extensive experiments on the KITTI and nuScenes datasets (Table~\ref{tab:kitti_nuscenes_comparison}) \cite{kitti,nuScenes}. The evaluation metrics follow standard protocols \cite{p2b,M2Track} and report Success and Precision under the One-Pass Evaluation (OPE) setting \cite{otb2013,kristan2016novel}. Specifically, the nuScenes dataset presents additional challenges compared to KITTI due to higher scene complexity and larger-scale annotations. 

\begin{table}[H]
\vspace{-2mm}
    \centering
    \caption{Comparison of KITTI and nuScenes dataset statistics.}
    \normalsize
    \begin{adjustbox}{max width=0.6\linewidth}
        \begin{tabular}{l|c|c}
            \toprule
            Attribute & KITTI~\cite{kitti} & nuScenes~\cite{nuScenes} \\
            \midrule
            Sequences/Scenes     & 50       & 1k \\
            LiDAR Frames         & 19k      & 390k \\
            Annotated Frames     & 8k       & 40k \\
            Classes              & 8        & 23 \\
            Data Size            & 35 GB    & $\sim$500 GB \\
            \bottomrule
        \end{tabular}
    \end{adjustbox}
    \label{tab:kitti_nuscenes_comparison}
    \vspace{-2mm}
\end{table}
\subsection{Experimental Results}

\begin{table*}[t]
\vspace{-1mm}
    \centering
    \caption{Comparisons with state-of-the-art methods on the nuScenes dataset. \textit{Success} / \textit{Precision} are used for evaluation. \textbf{Bold} denotes the best result. Mean is frame-weighted.}

    \resizebox{0.9\textwidth}{!}{
    \normalsize
    \begin{tabular}{c|cc|ccccc|c} \toprule
          Paradigm & Tracker & Source & Car [64,159] & Pedestrian [33,227] & Truck [13,587] & Trailer [3,352] & Bus [2,953] & Mean [117,278] \\ \midrule
          \multirow{13}{*}{Template-based}  & SC3D~\cite{sc3d} & CVPR'19 & 22.31 / 21.93 & 11.29 / 12.65 & 30.67 / 27.73 & 35.28 / 28.12 & 29.35 / 24.08 & 20.70 / 20.20 \\
                                            & P2B~\cite{p2b} & CVPR'20 & 38.81 / 43.18 & 28.39 / 52.24 & 42.95 / 41.59 & 48.96 / 40.05 & 32.95 / 27.41 & 36.48 / 45.08 \\
                                            & PTT~\cite{ptt-journal} & TMM'23 & 41.22 / 45.26 & 19.33 / 32.03 & 50.23 / 48.56 & 51.70 / 46.50 & 39.40 / 36.70 & 36.33 / 41.72 \\
                                            & BAT~\cite{bat} & ICCV'21 & 40.73 / 43.29 & 28.83 / 53.32 & 45.34 / 42.58 & 52.59 / 44.89 & 35.44 / 28.01 & 38.10 / 45.71 \\
          % &V2B~\cite{v2b} & NIPS'21 & 54.40 / 59.70 & 30.10 / 55.40 & 53.70 / 54.50 & 54.90 / 51.44 & - / - & - / - \\
                                            & PTTR~\cite{pttr} & CVPR'22 & 51.89 / 58.61 & 29.90 / 45.09 & 45.30 / 44.74 & 45.87 / 38.36 & 43.14 / 37.74 & 44.50 / 52.07 \\
                                            & SMAT~\cite{smat} & RAL'22 & 43.51 / 49.04 & 32.27 / 60.28 & - / - & - / - & 39.42 / 34.32 & - / - \\
                                            & STTracker~\cite{cui2023sttracker} & RAL'23 & 56.11 / 69.07 & 37.58 / 68.36 & 54.29 / 60.71 & 36.31 / 36.07 & 48.13 / 55.48 & 49.88 / 66.61 \\
                                            & GLT-T~\cite{glt} & AAAI'23 & 48.52 / 54.29 & 31.74 / 56.49 & 52.74 / 51.43 & 57.60 / 52.01 & 44.55 / 40.69 & 44.42 / 54.33 \\        
                                            & FlowTrack~\cite{li2024flowtrack} & IROS'24 & 60.29 / 71.07 & 37.60 / 67.64 & - / - & 55.39 / 62.70 & - / - & - / - \\           
                                            & M$^2$Track~\cite{M2Track} & CVPR'22 & 55.85 / 65.09 & 32.10 / 60.92 & 57.36 / 59.54 & 57.61 / 58.26 & 51.39 / 51.44 & 49.23 / 62.73 \\
                                            & MBPTrack~\cite{mbptrack} & ICCV'23 & 62.47 / 70.41 & 45.32 / 74.03 & 62.18 / 63.31 & 65.14 / 61.33 & 55.41 / 51.76 & 57.48 / 69.88 \\
                                            & MoCUT~\cite{cutrack} & ICLR'24 & 57.32 / 66.01 & 33.47 / 63.12 & 61.75 / 64.38 & 60.90 / 61.84 & 57.39 / 56.07 & 51.19 / 64.63 \\
                                        & 
                                        PillarTrack~\cite{pillartrack} & ArXiv'24 & 47.12 / 57.72 & 34.18 / 64.93 & 54.82 / 54.41 & 57.70 / 54.63 & 44.68 / 40.73 & 44.59 / 58.86 \\
                                        &
                                        BEVTrack~\cite{bevtrack} & IJCAI'25 & 64.31 / 71.14 & \textbf{46.28} / \textbf{76.77} & 66.83 / 67.04 & \textbf{74.54} / \textbf{71.62} & \textbf{61.09} / 56.68 & \textbf{59.71} / 71.19 \\ &
                                            TrackAny3D~\cite{wang2025trackany3d} & ICCV'25 & 59.30 / 66.46 & 40.37 / 68.70 & 62.70 / 62.80 & 66.12 / 59.20 & 61.01 / \textbf{58.02} & 54.57 / 66.25 \\ \midrule
          
          \rowcolor{gray!18}\multirow{3}{*}{Template-free} & \textbf{TFTrack-Voxel} & \textbf{Ours} & \textbf{64.59} / \textbf{72.41} & 43.94 / 72.33 & 66.53 / 68.48 & 72.48 / 69.00 & 57.54 / 55.59 & 59.01 / \textbf{71.41} \\
          \rowcolor{gray!18}                              & \textbf{TFTrack-Point} & \textbf{Ours} & 64.58 / 72.13 & 38.53 / 64.28 & \textbf{67.98} / \textbf{70.30} & 72.55 / 70.03 & 56.13 / 54.17 & 57.61 / 69.12 \\
          \rowcolor{gray!18}                              & \textbf{TFTrack-Pillar} & \textbf{Ours} & 63.18 / 71.56 & 43.07 / 72.40 & 67.27 / 69.70 & 70.17 / 68.12 & 57.95 / 56.92 & 58.01 / 71.10 \\ \bottomrule
    \end{tabular}%
    }%
    \label{tab:nusenes_results}
    \vspace{-3mm}
\end{table*}

\begin{table*}[h]
\centering
\caption{Comparison of \textbf{Template-Free} vs. \textbf{Template-Based} with Voxel view on nuScenes dataset (Success / Precision). Mean is frame-weighted.
}
\resizebox{0.8\linewidth}{!}{
\begin{tabular}{c|c|ccccc|c}
\toprule
\textbf{Method} & FLOPs (G) & Car & Ped. & Truck & Trailer & Bus & Mean \\
\midrule
\rowcolor{gray!18} Template-Free-Voxel & 
\textbf{34.7} & 
\textbf{64.6} / \textbf{72.4} & 
43.9 / 72.3 & 
\textbf{66.5} / \textbf{68.5} & 
\textbf{72.5} / \textbf{69.0} & 
\textbf{57.5} / \textbf{55.6} & 
59.0 / 71.4\\
Template-Based-Voxel & 
67.1 & 
64.6 / 72.0 & 
\textbf{45.6} / \textbf{74.2} & 
64.4 / 65.3 & 
70.2 / 66.1 & 
58.5 / 56.1 & 
59.2 / 71.2 \\
\bottomrule
\end{tabular}
}
\label{tab:free_based_compare}
\vspace{-3mm}
\end{table*}

\begin{table*}[h]
\centering
\caption{Backbone evaluation on nuScenes dataset. \textbf{Bold} denotes the best result. Mean is frame-weighted.}
\resizebox{0.8\linewidth}{!}{
\normalsize
\begin{tabular}{c|c|cccccc}
      \toprule[0.4mm]
      Method & Version & Car & Pedestrian & Truck & Trailer & Bus & Mean \\
      \midrule

      \multirow{2}{*}{TFTrack-Voxel}
      & VoxelNet~\cite{voxelnet} & \textbf{64.6}/\textbf{72.4} & \textbf{43.9}/\textbf{72.3} & \textbf{66.5}/\textbf{68.5} & \textbf{72.5}/\textbf{69.0} & \textbf{57.5}/\textbf{55.6} & \textbf{59.0}/\textbf{71.4} \\
      & FocalsConv~\cite{chen2022focal} & 60.7/70.1 & 40.7/68.2 & 62.9/64.5 & 68.4/62.8 & 52.6/51.9 & 55.3/68.2 \\
      \midrule
      \multirow{2}{*}{TFTrack-Point}
      & Point-MAE \cite{pointmae} & 64.3/72.1 & \textbf{38.5}/64.2 & 66.7/69.8 & 71.3/69.5 & 55.6/53.9 & 57.2/69.0 \\
      & PTv3 \cite{ptv3} & \textbf{64.6}/\textbf{72.1} & \textbf{38.5}/\textbf{64.3} & \textbf{68.0}/\textbf{70.3} & \textbf{72.6}/\textbf{70.0} & \textbf{56.1}/\textbf{54.2} & \textbf{57.6}/\textbf{69.1} \\
      \midrule
      \multirow{2}{*}{TFTrack-Pillar}
      & PillarNet~\cite{shi2022pillarnet} & \textbf{63.2}/\textbf{71.6} & \textbf{43.1}/\textbf{72.4} & \textbf{67.3}/\textbf{69.7} & \textbf{70.2}/\textbf{68.1} & \textbf{58.0}/\textbf{56.9} & \textbf{58.0}/\textbf{71.1} \\
      & PointPillars~\cite{pointpillars} & 59.8/68.3 & 40.2/69.5 & 63.4/66.1 & 66.8/65.0 & 54.9/53.7 & 54.7/67.9 \\
      \bottomrule[0.4mm] 
\end{tabular}}
% \vspace{-1mm}
\vspace{-5mm}
\label{table_backbone}
\end{table*}

\subsubsection{\textbf{Results on KITTI dataset}}
We report TFTrack results on the widely adopted but limited-scale KITTI dataset. As shown in Table~\ref{tab:kitti_results}, TFTrack-Voxel, TFTrack-Point and TFTrack-Pillar achieve mean Success/Precision scores of 64.0/81.6, 64.4/83.7, and 60.8/79.0, respectively. Meanwhile, TFTrack achieves strong efficiency at 92/121/95 FPS on RTX 3090, with TFTrack-Voxel further reaching 10.87 ms end-to-end latency and 185/226 MB peak allocated/reserved GPU memory. Although BEVTrack achieves higher KITTI accuracy, TFTrack offers a substantially simpler single-frame design. As summarized in Table~\ref{tab:kitti_nuscenes_comparison}, KITTI is substantially smaller and less diverse than nuScenes, making it more prone to overfitting. Thus, strong KITTI performance may not fully reflect generalization to diverse large-scale environments, motivating further evaluation on nuScenes.

\subsubsection{\textbf{Results on nuScenes dataset}}
We evaluated TFTrack on the large-scale nuScenes dataset, known for its sparse point clouds and complex scenes. As shown in Table~\ref{tab:nusenes_results}, our three variants, TFTrack-Voxel, TFTrack-Point, and TFTrack-Pillar, achieve highly competitive performance across all object categories. Notably, TFTrack-Voxel achieves 59.01/71.41 mean Success/Precision, remaining competitive with BEVTrack (59.71/71.19). TFTrack-Point achieves the best Truck and Trailer results among our variants, highlighting its strong capacity for modeling large vehicles. TFTrack-Pillar also yields solid results with a favorable balance of accuracy and efficiency. These results validate the effectiveness of our single-frame, template-free design and its ability to generalize well to dynamic and challenging tracking scenarios without dual-frame matching.

% \vspace{-1mm}

\subsubsection{\textbf{Template-Free vs.\ Template-Based}}
Table~\ref{tab:free_based_compare} compares our template-free \textit{TFTrack-Voxel} against a dual-frame \textit{Template-Based-Voxel} that processes both ${P}_{t-1}$ and ${P}_t$. Despite halving the FLOPs (34.7\,G vs.\ 67.1\,G), the template-free variant approaches the dual-frame mean Success/Precision (59.0/71.4 vs.\ 59.2/71.2), suggesting that the explicit template branch adds redundant computation.

\subsubsection{\textbf{Backbone Analysis}}
In Table~\ref{table_backbone}, we benchmark three sparse backbones on nuScenes: Point-MAE vs. PTv3 \cite{pointmae,ptv3} for \textit{TFTrack-Point}, VoxelNet vs. FocalsConv \cite{voxelnet,chen2022focal} for \textit{TFTrack-Voxel}, and PillarNet vs. PointPillars \cite{shi2022pillarnet,pointpillars} for \textit{TFTrack-Pillar}. Upgrading to PTv3 yields a small gain (57.5/69.1 vs. 57.2/69.0), VoxelNet improves by +3.7/+3.2 over FocalsConv (59.0/71.4 vs. 55.3/68.2), and PillarNet outperforms PointPillars by +3.3/+3.2 (58.0/71.1 vs. 54.7/67.9). These results demonstrate the flexibility of TFTrack across different backbones.

\begin{table*}[t]
\centering
\caption{Ablation study of re-centering and Box-Conditioned Feature Encoding in \textbf{TFTrack-Voxel} on nuScenes dataset \cite{nuScenes}. "Re-center" denotes the coordinate transformation of the current frame using the previous box center. "BCFE" denotes Box-Conditioned Feature Encoding. Mean is frame-weighted.}
\label{tab:ablation}
\setlength{\tabcolsep}{4.2pt}      
\renewcommand{\arraystretch}{1.08} 
\footnotesize                      

\begin{tabular*}{\textwidth}{@{\extracolsep{\fill}} c|c|cc|ccccc|c}
\toprule[0.4mm]
Method & \# & Re-center & BCFE & Car & Pedestrian & Truck & Trailer & Bus & Mean \\
\midrule
\multirow{4}{*}{TF}
& \textcircled{1} & \checkmark & \checkmark & 64.6/72.4 & 43.9/72.3 & 66.5/68.5 & 72.5/69.0 & 57.5/55.6 & 59.0/71.4 \\
& \textcircled{2} & \checkmark &            & 61.1/69.2 & 40.2/68.7 & 61.1/62.5 & 66.9/61.8 & 52.4/50.3 & 55.1/67.6\\
& \textcircled{3} &            & \checkmark  & 55.8/62.8 & 36.5/64.8 & 58.2/59.0 & 63.5/60.1 & 49.1/47.2 & 50.7/62.5 \\
& \textcircled{4} &            &             & 55.1/62.7 & 35.9/65.4 & 58.8/58.3 & 64.1/59.7 & 48.7/46.9 & 50.2/62.5 \\
\bottomrule[0.4mm]
\end{tabular*}
\vspace{-4mm}
\end{table*}

\begin{table}[ht]
    \centering
    \caption{Effect of template masking on \textbf{Template-Based-Voxel} (\textbf{TB}) under different masking ratios on nuScenes (Success / Precision).}
    
    \begin{adjustbox}{max width=\linewidth}
        \begin{tabular}{c|c|cc}
            \toprule
            Method & Mask Ratio & Car & Pedestrian \\
            \midrule
            \multirow{6}{*}{TB} 
            & 0\% (Full Template)   & 64.6 / 72.0 & 45.6 / 74.2 \\
            & 20\% Masked           & 64.5 / 71.8 & 44.8 / 73.7 \\
            & 40\% Masked           & 64.2 / 71.1 & 44.2 / 73.6 \\
            & 60\% Masked           & 63.8 / 71.2 & 43.2 / 73.2 \\
            & 80\% Masked           & 62.7 / 70.3 & 41.3 / 70.2 \\
            & 100\% Masked          & 61.9 / 69.8 & 39.8 / 68.1 \\
            \bottomrule
        \end{tabular}
    \end{adjustbox}
    \label{tab:masking}
\end{table}

\begin{table}[t]
\centering
\caption{Tracklet-level robustness under challenging conditions on nuScenes Car (Success / Precision). ``Late 25\%'' denotes the last quarter of each tracklet; ``Sp. init.'' denotes fewer than $10$ target points in the first frame; ``Fast start'' denotes initial GT speed larger than $10$ m/s.}\label{tab:tracklet_robust}
\scriptsize
\setlength{\tabcolsep}{1.6pt}
\renewcommand{\arraystretch}{1.0}
\resizebox{\columnwidth}{!}{%
\begin{tabular}{lcccc}
\toprule
Method & Mean & Late 25\% & Sp. init. & Fast start \\
\midrule
M$^2$Track & 60.81 / 68.73 & 45.38 / 53.77 & 51.23 / 60.46 & 14.38 / 13.05 \\
MBPTrack   & 65.72 / 73.06 & 51.56 / 60.14 & 57.98 / 66.09 & 12.59 / 11.50 \\
BEVTrack   & 67.72 / 73.77 & 52.83 / 60.98 & \textbf{62.37} / 69.16 & 9.75 / 7.79 \\
TFTrack    & \textbf{67.79} / \textbf{74.46} & \textbf{54.27} / \textbf{63.16} & 61.76 / \textbf{69.45} & \textbf{15.19} / \textbf{14.17} \\
\bottomrule
\end{tabular}%

}
\vspace{-5mm}

\end{table}

% \subsubsection{\textbf{Ablation of Proposed Module}}
% In Table~\ref{tab:ablation}, we isolate the two core modules in \textit{TFTrack-Voxel}. Target-Centric transformation (Re-centering) alone (coordinate transform using ${B}_{t-1}$) raises mean Success/Precision from 53.3/59.1 to 56.3/62.3; Box-Conditioned Feature Encoding (BCFE) alone yields 52.8/58.6; combining both recovers the full 60.5/67.0 gain. These results demonstrate simple re-centering and BCFE are each effective and complementary for single-frame tracking.

\subsubsection{\textbf{Ablation of Proposed Module}}
In Table~\ref{tab:ablation}, we isolate the two core modules in \textit{TFTrack-Voxel}. Re-centering alone raises mean Success/Precision from 50.2/62.5 to 55.1/67.6 by reducing translation variance and establishing a target-centric reference frame around ${B}_{t-1}$. BCFE alone yields a marginal gain to 50.7/62.5, as the size prior is less effective in the global coordinate system. When combined with Re-centering, BCFE further improves performance to 59.0/71.4. These results show that Re-centering provides motion-aware coordinate normalization, while BCFE supplies complementary object-scale conditioning.

\subsubsection{\textbf{Template Masking}}
In Table~\ref{tab:masking}, we progressively mask the template input in the dual-frame setup (0-100\% for two main categories test, where masking is performed by replacing template points with zeros. Masking up to 80\% causes only minor drops. This indicates that our implicit geometric priors capture the bulk of tracking cues without relying on full templates.

% \subsubsection{\textbf{Robustness for Sparsity.}} We evaluate TFTrack under challenging conditions. In sparse cases (initial frame $<$10 points), TFTrack achieves 59.7\%/67.6\%, outperforming M$^2$Track \cite{M2Track} (48.0\%/56.4\%). For high-speed scenes (acceleration $>$10 m/s$^2$), it still maintains 5.48\% success. Moreover, selectively adding a template in sparse cases brings gains (59.71\% $\rightarrow$ 60.28\%), validating the orthogonality of template usage without altering the core design.

\subsubsection{\textbf{Robustness Analysis}}
We further evaluate TFTrack with tracklet-level diagnostics on nuScenes Car under challenging conditions, as shown in Table~\ref{tab:tracklet_robust}.  We report the overall tracklet mean, late-stage performance on the last 25\% of each tracklet to measure drift accumulation, sparse-initial cases to evaluate weak geometric observations, and fast-start cases to examine large initial motion. TFTrack achieves the best Precision in all settings and the best Success on the overall, late-stage, and fast-start subsets. These results show that the proposed previous-box prior remains robust under common short-term tracking challenges while preserving the template-free design.

\subsubsection{\textbf{Visualization}}
\begin{figure}[ht]
    \centering
    \includegraphics[width=0.7\linewidth]{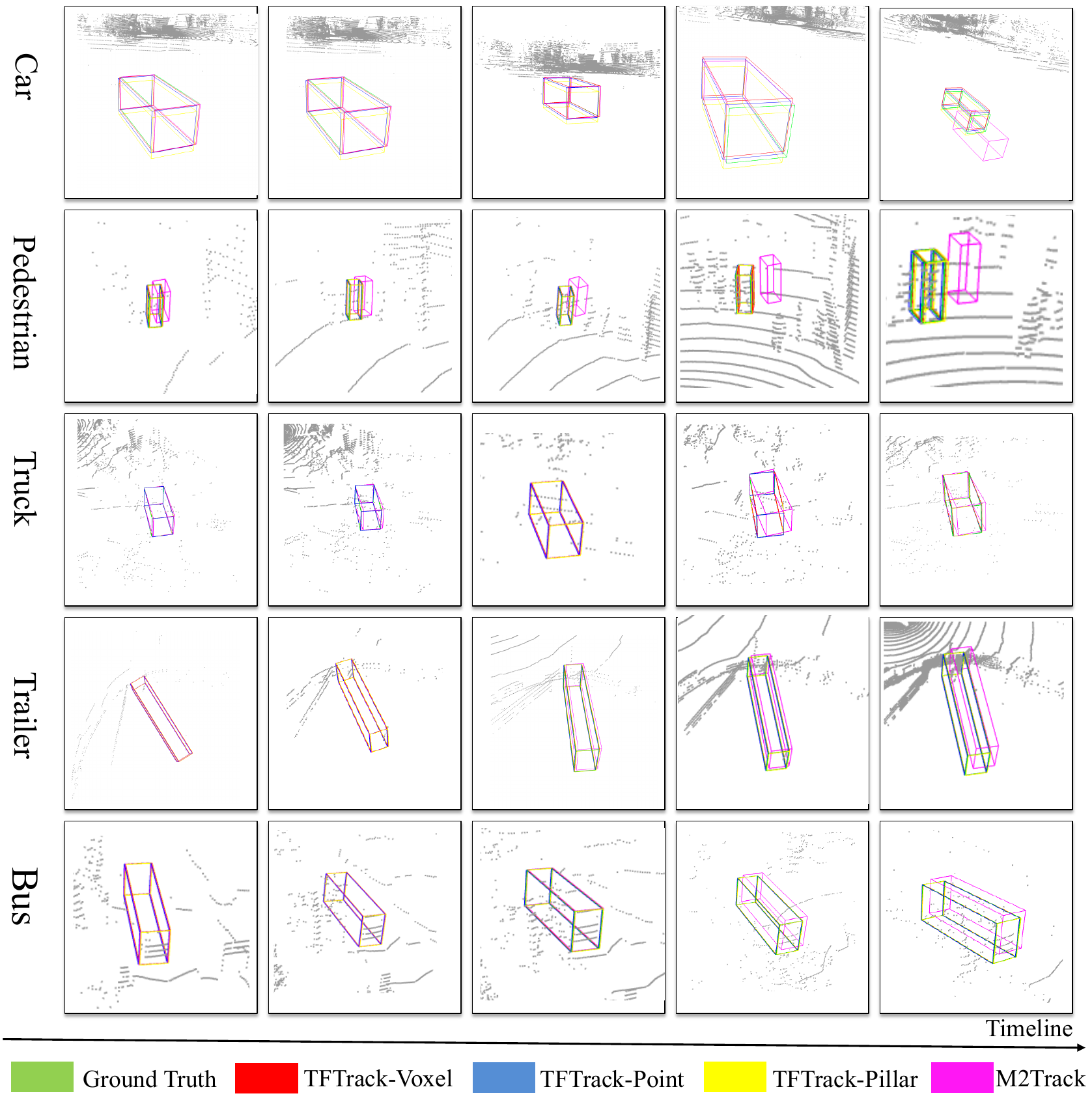}
    \vspace{-2mm}
    \caption{Visualization of tracking results on nuScenes dataset.}
    \label{fig:results}
    \vspace{-2mm}
\end{figure}

Fig.~\ref{fig:results} illustrates tracking results on the nuScenes dataset. We compare three Template‐free variants, TFTrack–Voxel (red), TFTrack–Point (blue), and TFTrack–Pillar (yellow), against the motion‐centric M$^2$Track (magenta) and ground truth (green). Across diverse object classes (Car, Pedestrian, Truck, Trailer, Bus), TFTrack consistently aligns closely with ground truth, showing robust tracking performance without a template point-cloud input.
\vspace{-2mm}

\section{CONCLUSION}
\vspace{-1mm}
\label{sec:conclusion}
This paper presents \textbf{TFTrack}, the first template-free and single-frame method for 3D SOT. Departing from motion-centric dual-frame paradigms, TFTrack leverages a simple yet powerful insight: the previous bounding box center and box size provide sufficient geometric prior for motion estimation from the current frame alone. By reformulating 3D SOT as a single-frame transformation problem, TFTrack eliminates redundant template encoding and cross-frame feature correlation, reducing computation and latency. We instantiate the framework with three views, point, pillar, and voxel, to explore trade-offs between accuracy and efficiency. Extensive experiments show that TFTrack achieves competitive accuracy with high inference efficiency, making it well-suited for onboard deployment in embedded systems.

\noindent\textbf{Limitations}: TFTrack may degrade under extremely sparse LiDAR observations, severe occlusion, or adverse weather. Following standard 3D SOT benchmarks, the current implementation predicts 4-DoF rigid motion with constant object size, which limits handling of strong deformation. Future work will explore multimodal sensing, lightweight recovery mechanisms, and richer motion representations.

\vspace{-2mm}
% \clearpage
\bibliographystyle{IEEEtran}
\bibliography{IEEEabrv,strings,refs}

\end{document}